\documentclass[letterpaper,10pt,conference]{ieeeconf}
\IEEEoverridecommandlockouts
\usepackage{amsmath,amssymb,booktabs,multirow,graphicx,xspace,cite,url,microtype,balance}

\newcommand{\method}{ExecutionBridge\xspace}
\newcommand{\wote}{WoTE\xspace}

\newcommand{\pdmsgain}[1]{%
    \rlap{\raisebox{0.55ex}{\scriptsize\,(+#1)}}%
}

\title{Plans Are Not Executions: Adapting Frozen World-Model Planners to Controller-Induced Execution Shift}

\author{%
Danqi Zhao$^{1}$, Keyu Chen$^{1}$, Wenchao Sun$^{1}$,
Minglei Huang$^{3}$, Jianhe Zhang$^{2}$,\\
Yan Zhu$^{2}$, Hao Cheng$^{1}$, Zifan Cao$^{1}$,
and Sifa Zheng$^{1,*}$%
\thanks{$^{1}$School of Vehicle and Mobility,
Tsinghua University, Beijing, China.
E-mail: \texttt{zdq21@mails.tsinghua.edu.cn}.
$^{*}$Corresponding author: Sifa Zheng
(\texttt{zsf@tsinghua.edu.cn}).}%
\thanks{$^{2}$DENSO (China) Investment Co., Ltd.,
China.}%
\thanks{$^{3}$Shanghai Technical Center,
DENSO (China) Investment Co., Ltd.,
Shanghai, China.}%
}

\begin{document}
\maketitle
\begin{abstract}
World-model planners evaluate candidate trajectories by predicting their future consequences, implicitly relying on the plan-to-motion relationship learned during training. At deployment, a different downstream controller may execute the same planned trajectory differently, creating a planning–execution gap between the frozen model’s imagined future and actual execution.
A natural remedy is to replace the planned trajectory with its predicted or observed execution. However, we find that better execution fidelity does not necessarily yield better decisions:
% even exact execution, which eliminates geometric execution error in our diagnostic experiment, changes downstream PDMS by only $0.277$. 
directly substituting predicted execution trajectories yields only a 0.048 PDMS improvement across 10 unseen controller configurations, far smaller than the gain from our proposed adaptation.
This suggests a semantic mismatch between a planned trajectory, which serves as a reference for downstream control, and an executed trajectory, which already describes the resulting post-controller motion.

We introduce \textbf{ExecutionBridge}, which infers controller-specific execution characteristics from planned--executed trajectory pairs and uses them to adapt the native ego latent rollout while keeping the original planned trajectory unchanged. The planner, world model, and reward head remain frozen. On NAVSIM, ExecutionBridge consistently improves planning across
unseen controller configurations and multiple frozen world-model
planners, with substantially larger gains in challenging driving
scenes where execution mismatch more strongly affects candidate
evaluation. 

\end{abstract}

\section{Introduction}
\label{sec:intro}

\begin{figure}[t]
    \centering
    \includegraphics[width=\columnwidth]
    {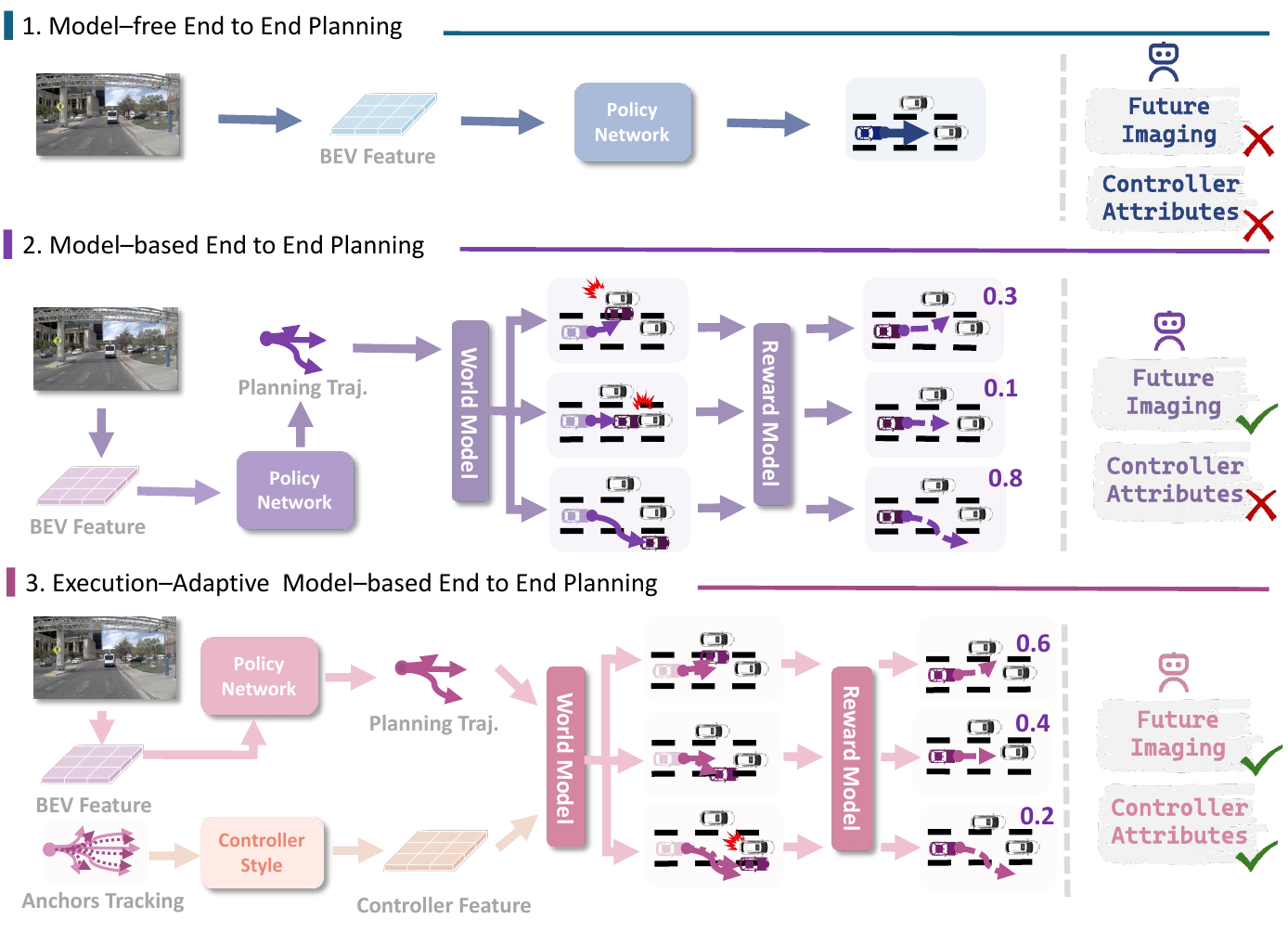}
    \caption{
    \textbf{Evolution of end-to-end planning paradigms.}
    Model-free planners predict or score trajectories directly from the
    current scene, without explicitly modeling their future consequences.
    Model-based planners additionally use a learned world model to imagine
    candidate-conditioned futures for trajectory evaluation.
    Conventional model-based planning typically assumes that candidate
    trajectories are executed according to the plan-to-motion relationship
    learned during training.
    Execution-adaptive model-based planning further incorporates
    controller-specific information, allowing candidate futures
    to be evaluated under deployment-specific execution.
    }
    \label{fig:INTRO-1}
\end{figure}

End-to-end autonomous driving~\cite{Chen2024E2E,li2025law} has evolved from directly predicting a
single planning output~\cite{Hu2022STP3,Hu2023PlanningOriented,Jiang2023VAD}
toward generating and evaluating multiple candidate
trajectories~\cite{Chen2024VADV2,Li2024HydraMDP}.
As illustrated in Fig.~\ref{fig:INTRO-1}, existing planning paradigms can be
broadly divided into model-free and model-based approaches.
Model-free planners predict or score trajectories primarily from the current
scene representation, without explicitly reasoning about their downstream
consequences.
In contrast, model-based planners use a learned world model to imagine the
future induced by each candidate before selecting a
trajectory~\cite{Wang2024DrivingFuture,Zheng2024OccWorld,
li2025wote,Zheng2025World4Drive}.
Such models represent future states in image, occupancy, BEV, or latent spaces,
enabling the planner to ask a seemingly simple but powerful question:
\emph{what will happen if this trajectory is executed?}

The answer, however, depends not only on the planned trajectory, but also on
how it is executed by the downstream controller.
World-model-based evaluation implicitly relies on the plan-to-motion
relationship learned during training.
At deployment, changes in feedback controllers, control gains, actuation
constraints, or vehicle dynamics can cause the same planned trajectory to
produce different ego motion~\cite{Paden2016Survey,Rajamani2006Vehicle,
Mayne2000MPC}.
A frozen world model may therefore predict a future that is consistent
with its implicit training-time plan-to-motion mapping, yet inconsistent with
the execution actually encountered by the deployed vehicle.
We refer to this discrepancy as the \emph{planning--execution gap}.

A natural remedy is to predict how the deployment controller will execute each
candidate trajectory and use the resulting executed trajectory in place of the
original plan.
Surprisingly, we find that better execution fidelity does not necessarily
translate into better planning decisions.
Even when the executed trajectory is known exactly in simulation, direct
substitution yields only limited improvements in candidate ranking and
downstream planning quality.

We attribute this failure to a semantic mismatch between planned and executed
trajectories.
Although both are represented as trajectories, they play fundamentally
different roles.
During training, the world model is conditioned on a \emph{planned trajectory}
and implicitly learns how such an input relates to ego motion and subsequent
scene evolution.
An \emph{executed trajectory}, in contrast, already describes the motion
resulting after downstream control.
Substituting it as the trajectory condition therefore asks the frozen world
model to reinterpret post-controller motion through its implicit
plan-to-motion mapping.
As a result, a geometrically more accurate execution does not necessarily
produce a more accurate imagined consequence or a better candidate ranking.

\begin{figure*}[t]
    \centering
    \includegraphics[width=0.98\textwidth]{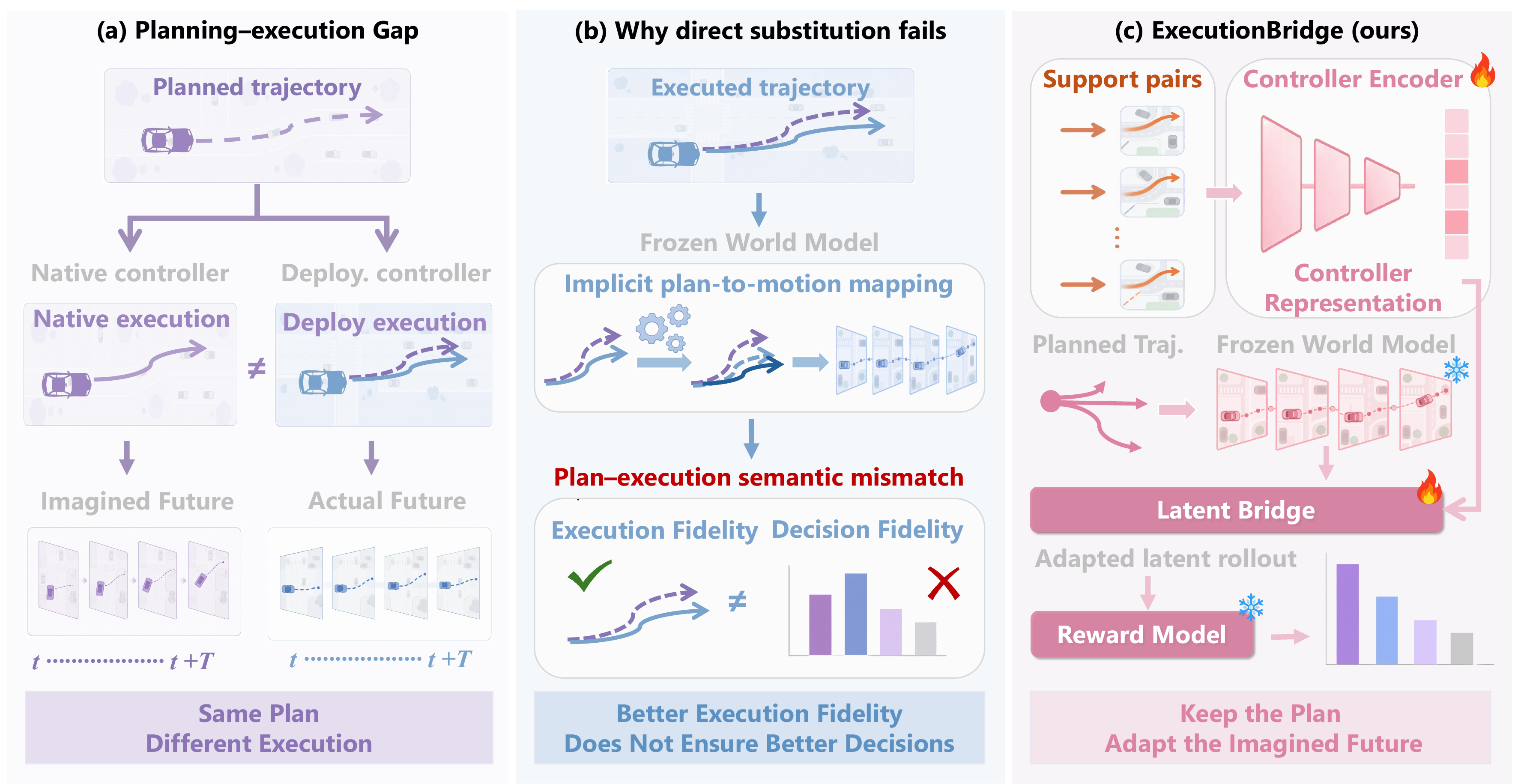}
    \caption{
    \textbf{Planning--execution gap and motivation for ExecutionBridge.}
    \textbf{(a)} The same planned trajectory can lead to different executions
    at deployment from those implicitly assumed by the frozen world model,
    causing its imagined future to deviate from the future actually encountered.
    \textbf{(b)} Directly substituting the planned trajectory with its executed
    counterpart improves execution fidelity, but introduces a
    \emph{plan--execution semantic mismatch}: the executed trajectory can still
    be interpreted through the model's implicit plan-to-motion mapping, and
    therefore does not necessarily improve candidate ranking.
    \textbf{(c)} ExecutionBridge instead retains the original planned trajectory
    as the conditioning input and uses controller-specific execution evidence
    to adapt the imagined future used for candidate evaluation.
    }
    \label{fig:overview}
\end{figure*}

This observation suggests a different adaptation principle:
execution information should adapt how the future is imagined, rather than
replace the planned trajectory itself.
As illustrated in Fig.~\ref{fig:overview}, we retain the original planned
trajectory as the conditioning input and use controller-specific execution
evidence to adapt the future representation.
% used for candidate evaluation.

Based on this principle, we propose \textbf{ExecutionBridge}, a lightweight
adaptation framework for frozen world-model planners.
Given a support set of paired planned and executed trajectories from the
deployment controller, ExecutionBridge infers a compact controller
representation that captures persistent execution characteristics.
For each candidate, the frozen world model first produces its native latent
rollout from the original planned trajectory, after which ExecutionBridge
adapts the ego-related latent using the inferred controller representation.
The planner, world model, and reward head remain frozen, enabling
controller-conditioned candidate evaluation without replacing the planned
trajectory with post-controller motion.

Our contributions are threefold:
\begin{itemize}

\item We identify \emph{controller-induced execution shift} as a source of the
planning--execution gap in frozen world-model planning.
Through a diagnostic study, we show that even exact post-controller execution
provides only limited benefit when directly substituted for the planned
trajectory, demonstrating that execution fidelity does not necessarily imply
decision fidelity.

\item We introduce \textbf{ExecutionBridge}, which infers controller-specific
execution characteristics from a set of paired planned and executed
trajectories and uses them to adapt the native ego latent rollout while
retaining the original planned trajectory and all pretrained planning
components unchanged.

\item We evaluate ExecutionBridge under a strict controller-disjoint protocol
and across multiple frozen world-model planners.
ExecutionBridge consistently improves planning across ten unseen controller
configurations, with substantially larger gains in challenging driving scenes
where execution shift has a greater impact on candidate evaluation and
selection.

\end{itemize}

\section{Related Work}
\label{sec:related}

\subsection{World Models for Planning}

World models learn predictive dynamics that enable agents to reason about
future consequences before acting~\cite{sutton1991dyna,hafner2019learning,
hafner2019dream,janner2019when}.
In autonomous driving, BEV-, occupancy-, latent-, and video-based world models
have been developed to predict future scene evolution and support downstream
planning~\cite{Hu2022STP3,Zheng2024OccWorld,Wang2024DriveDreamer,
Zheng2025World4Drive}.
Trajectory-evaluation frameworks such as
\wote~\cite{li2025wote} further use imagined future states to score
multiple candidate trajectories, making world-model prediction directly
responsible for candidate ranking.

These methods generally rely on the plan-to-motion relationship captured
during training remaining valid at deployment.
In practice, however, a change in downstream controller can cause the same
planned trajectory to produce different ego motion.
We study this setting for frozen world-model planners, where the planner,
world model, and reward head remain unchanged while deployment-time execution
behavior differs from that implicitly captured during pretraining.

\subsection{Dynamics Identification and Adaptation}

System identification and context-conditioned adaptation have long been used
to handle dynamics variation.
CaDM~\cite{Lee2020CaDM}, UP-OSI~\cite{Yu2017UPOSI}, and
RMA~\cite{Kumar2021RMA} infer deployment-specific dynamics or latent context
from interaction history to condition prediction or control, while
A-RMA~\cite{Kumar2022ARMA} further considers adaptation with estimated
dynamics latents.

Related ideas have also been explored in autonomous driving.
AdaptiveDriver~\cite{Vasudevan2025AdaptiveWorldModels} models
behavior-controller parameters of surrounding agents,
DynaDreamer~\cite{Wang2026DynaDreamer} incorporates identifiable ego dynamics
for cross-chassis adaptation, and AdaWM~\cite{Wang2025AdaWM} selectively
adapts the dynamics model or planning policy under deployment shift.

Our setting differs in both the adaptation target and deployment constraint:
the planner, world model, and reward head remain frozen, while
planned--executed trajectory pairs are used to infer controller-specific
execution characteristics that adapt the native latent rollout for candidate
evaluation.

\subsection{Execution-Aware Planning and Predictive Conditioning}

Prior work has addressed plan--execution mismatch either at the planning
level or through explicit execution modeling.
Drag-aware trajectory generation~\cite{Zhang2024DragAware} incorporates
controller-dependent tracking costs into planning, while robot-factored world
models~\cite{Kim2026RobotFactored} separate commanded actions, physical
execution, and subsequent environment evolution.
These approaches modify the planning objective or explicitly model the
plan-to-execution transition.

In contrast, \method keeps the pretrained world model and its planned
trajectory condition unchanged, using execution observations only to infer
controller-specific context that adapts the latent future for candidate
evaluation.

%==============================
\section{Problem Formulation}
\label{sec:problem}

At time $t$, the planner observes a scene represented by a world-model
latent state $z_t$ and proposes $K$ candidate trajectories for subsequent evaluation
\begin{equation}
    \mathcal{A}_t = \{a_t^k\}_{k=1}^{K},
\end{equation}
where each $a_t^k\in\mathbb{R}^{H\times 3}$ specifies the planned ego
position and yaw over a horizon of $H$ steps.

Let $F_\theta$ denote the frozen world model and $Q_\omega$ the frozen
reward head used for candidate evaluation. For each candidate $a_t^k$,
the world model predicts a candidate-conditioned future representation
from the planned trajectory:
\begin{equation}
    z_{t+1}^{0,k}
    =
    F_\theta(z_t,a_t^k),
    \qquad
    s_0^k
    =
    Q_\omega(z_{t+1}^{0,k}).
    \label{eq:native_rollout}
\end{equation}
Here, the superscript $0$ denotes the \emph{native}, unadapted rollout
of the pretrained world model. The frozen evaluator selects
\begin{equation}
    \hat{k}_0 = \arg\max_{k} s_0^k.
    \label{eq:native_selection}
\end{equation}

Through pretraining, the frozen world model implicitly captures a
plan-to-motion mapping from planned trajectories to the ego motion and
future scene evolution observed in the training distribution.
We use this mapping only as a conceptual description of the execution
behavior embedded in $F_\theta$; it need not correspond to an explicit
or separately identifiable dynamics module.
At deployment, however, a downstream controller $c$ may execute the
same planned trajectory differently. We denote the resulting
controller-specific plan-to-execution mapping by
\begin{equation}
    \tau_t^{c,k}
    =
    \mathcal{E}_c(a_t^k;x_t),
    \label{eq:execution_map}
\end{equation}
where $x_t$ is the observable ego state and
$\tau_t^{c,k}\in\mathbb{R}^{H\times3}$ denotes the executed ego
trajectory for $a_t^k$ under controller $c$.

When the deployment mapping $\mathcal{E}_c$ differs from the implicit
plan-to-motion mapping captured during pretraining, the frozen world
model continues to evaluate $a_t^k$ according to execution behavior
that may no longer match the deployed system.
We refer to this discrepancy between the execution implicitly assumed
by the frozen evaluator and that induced by the deployment controller
as the \emph{planning--execution gap}.
Consequently, the future imagined for a candidate may differ from the
future actually encountered during its execution.

We assume access to a support set of $N$ planned--executed
trajectory pairs collected under controller $c$,
\begin{equation}
    \mathcal{S}_c
    =
    \{(a_i,\tau_i^c)\}_{i=1}^{N},
    \label{eq:support_set}
\end{equation}
where the support samples are collected independently of the current
candidate set $\mathcal{A}_t$.

%=====================
\begin{figure*}[t]
    \centering
    \includegraphics[width=\textwidth]{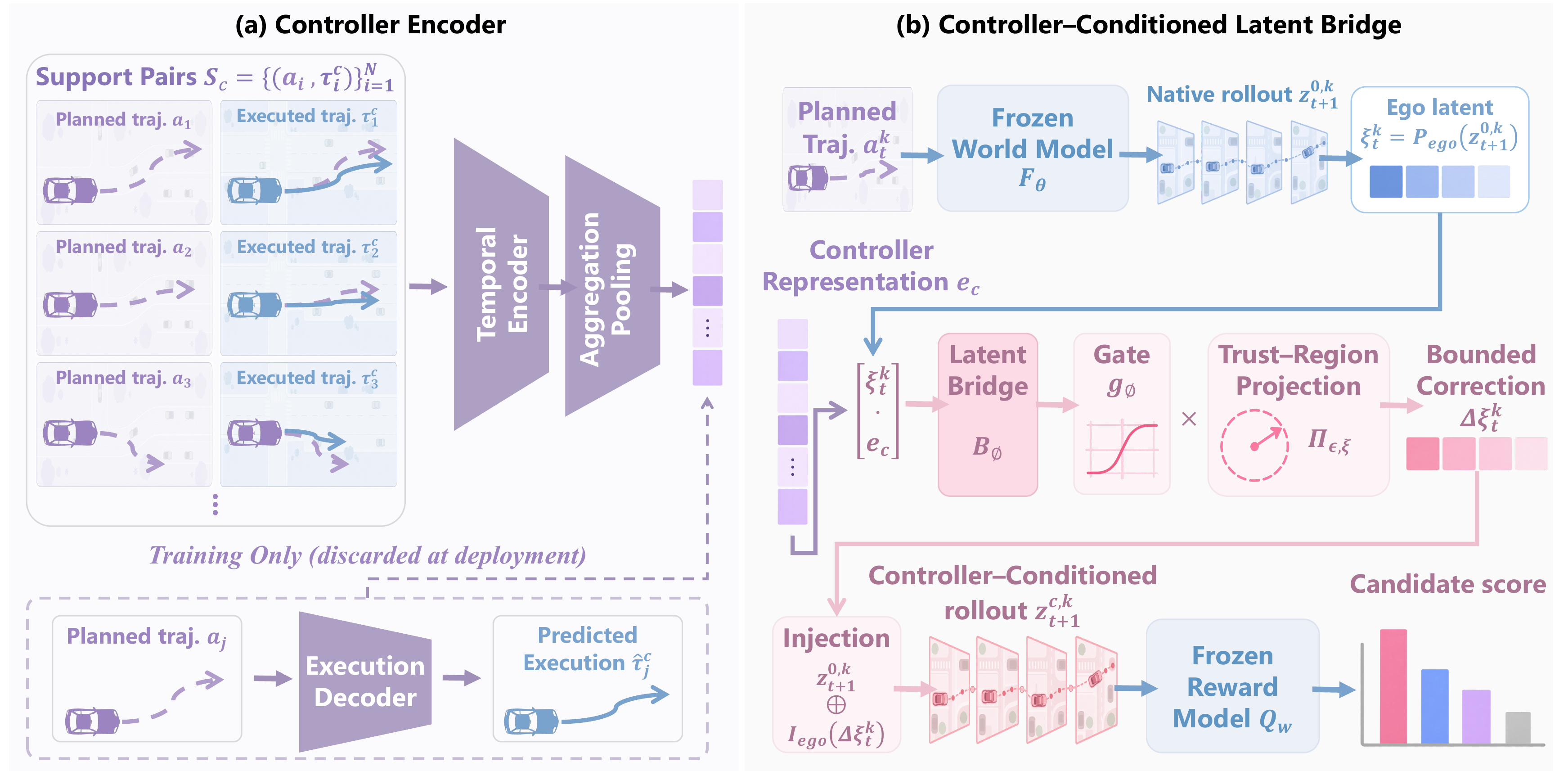}
    \caption{
    \textbf{Architecture of ExecutionBridge.}
    \textbf{(a) Controller Encoder.}
    Given a support set of planned--executed trajectory pairs
    $\mathcal{S}_c$, the Controller Encoder summarizes these observations into
    a compact controller representation $e_c$ that captures persistent
    controller-specific execution characteristics.
    An auxiliary execution decoder predicts the execution of a held-out query
    trajectory for training supervision and is discarded at deployment.
    \textbf{(b) Controller-Conditioned Latent Bridge.}
    For each candidate, the frozen world model first produces its native rollout
    from the original planned trajectory.
    Conditioned on $e_c$, the Latent Bridge predicts a gated correction to the
    ego latent, which is bounded by a trust-region projection before being
    injected into the rollout.
    The resulting controller-conditioned rollout is evaluated by the unchanged
    reward model for candidate scoring.
    }
    \label{fig:method_overview}
\end{figure*}

Given $\mathcal{S}_c$, our goal is to obtain controller-conditioned
candidate scores $\{\hat{s}_c^k\}_{k=1}^{K}$ without modifying the
candidate trajectories produced by the planner or any pretrained
planning component.
In particular, each original planned trajectory $a_t^k$ remains the
conditioning input to the frozen world model, while information inferred
from $\mathcal{S}_c$ is used only to adapt its candidate evaluation.
%========================

\section{Method}
\label{sec:method}
%====================================
%====================================
\subsection{Overview}
\label{sec}
\method addresses the planning--execution gap using a support set of
planned--executed trajectory pairs $\mathcal{S}_c$ collected from the
deployment controller.
Rather than replacing a planned trajectory with its executed counterpart,
\method retains the original trajectory as the conditioning input to the
frozen world model and adapts the resulting latent future using
controller-specific execution information.
Specifically, \method first obtains the native latent rollout from the frozen
world model and then applies a controller-conditioned correction to its
ego latent.

As illustrated in Fig.~\ref{fig:method_overview}, \method consists of two
lightweight components.
A \emph{Controller Encoder (CE)} summarizes $\mathcal{S}_c$ into a compact
controller representation $e_c$ that captures persistent controller-specific
execution characteristics.
A \emph{Latent Bridge (LB)} then conditions on $e_c$ to adapt the native ego
latent produced by the frozen world model.

For each candidate $a_t^k$, the frozen world model first produces its native
rollout from the original planned trajectory,
\begin{equation}
z_{t+1}^{0,k}=F_\theta(z_t,a_t^k).
\label{eq:method_native_rollout}
\end{equation}
\method then transforms this rollout into a controller-conditioned rollout
$z_{t+1}^{c,k}$ without modifying the planned trajectory.
The unchanged reward head evaluates
\begin{equation}
\hat{s}_c^k = Q_\omega(z_{t+1}^{c,k}),
\label{eq:adapted_score}
\end{equation}
and the final trajectory is selected as
\begin{equation}
\hat{k}_c = \arg\max_k \hat{s}_c^k.
\label{eq:adapted_selection}
\end{equation}
The planner, world model, and reward head remain frozen.

\subsection{Controller Representation}
\label{sec:controller_representation}

Given the support set
$\mathcal{S}_c=\{(a_i,\tau_i^c)\}_{i=1}^{N}$,
the Controller Encoder maps the support set \(\mathcal S_c\) to a controller representation
\begin{equation}
e_c = E_\eta(\mathcal{S}_c),
\label{eq:controller_embedding}
\end{equation}
where $e_c$ captures controller-specific execution characteristics.

%============相机图片=======
\begin{figure*}[t]
    \centering
    \includegraphics[width=\textwidth]
    {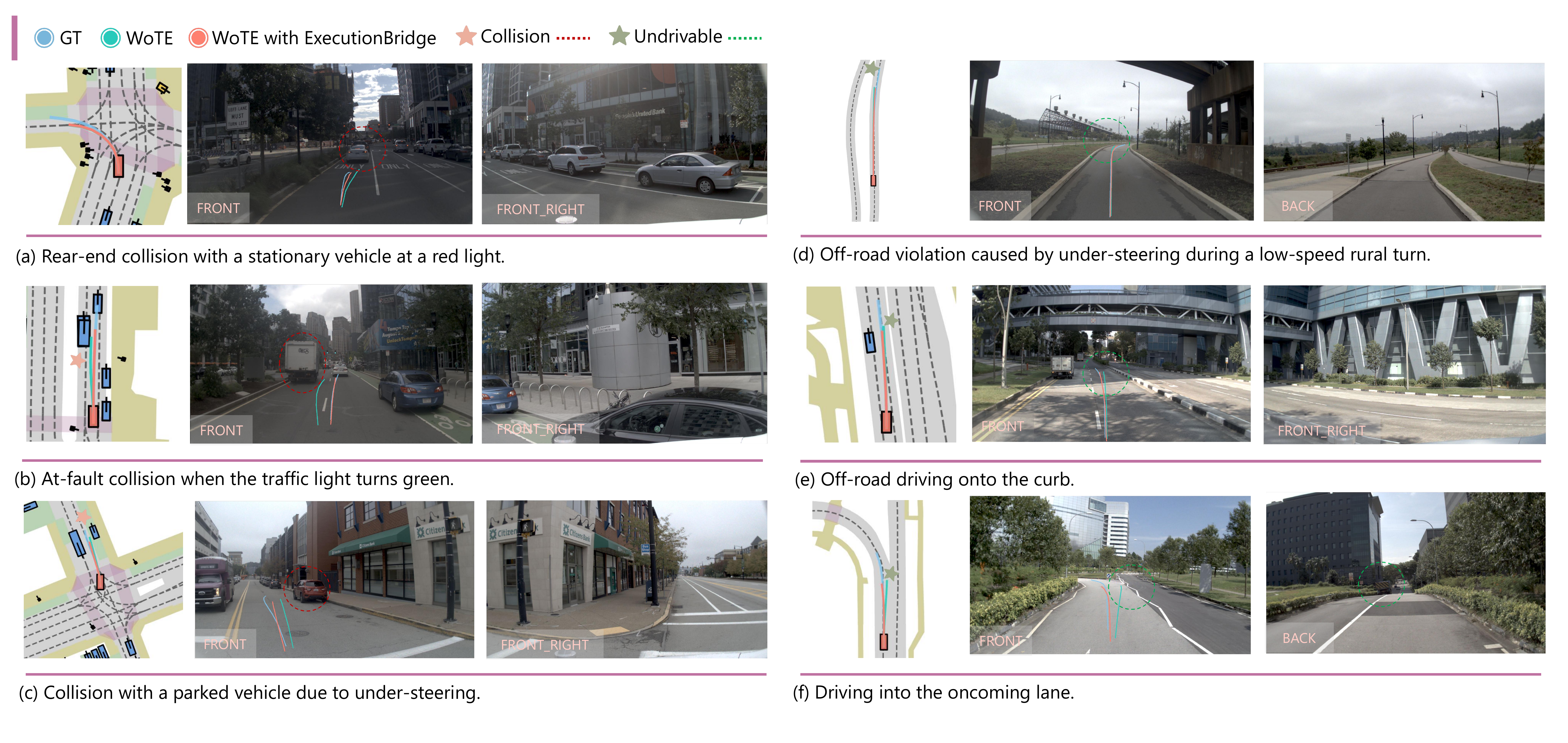}
    \caption{
    \textbf{Representative planning results under the standard NAVSIM setting.}
    Compared with frozen \wote, \method yields trajectory selections that are
    better aligned with controller-dependent execution in representative
    collision and drivable-area failure cases.
    }
    \label{fig:qualitative_standard}
\end{figure*}
%=============================

We learn this representation without controller labels or explicit dynamics
parameters.
Instead, we use an auxiliary support--query execution-prediction task.
Given an independently sampled trajectory $a_j$ with ego state $x_j$,
a training-only decoder predicts its execution conditioned on $e_c$:
\begin{equation}
\hat{\tau}_j^c
=
G_\psi(a_j,x_j,e_c).
\label{eq:execution_prediction}
\end{equation}

Support and query trajectories are sampled independently from the same
controller, and the encoder--decoder is optimized using
\begin{equation}
\mathcal{L}_{\mathrm{exec}}
=
\ell_{\mathrm{traj}}
\left(
\hat{\tau}_j^c,\tau_j^c
\right),
\label{eq:execution_loss}
\end{equation}
where $\ell_{\mathrm{traj}}$ measures trajectory position and heading errors.
Predicting the execution of unseen query trajectories encourages $e_c$ to
capture controller-specific execution characteristics that persist across
different planned trajectories, rather than memorizing individual support
samples.
Afterward, the Controller Encoder is frozen and the auxiliary decoder discarded.

%===============================
\subsection{Controller-Conditioned Latent Bridge}
\label{sec:latent_bridge}

Directly substituting an executed trajectory for the planned trajectory
changes the semantic role of the trajectory condition.
We therefore retain the original candidate $a_t^k$ as the conditioning input
to the frozen world model and adapt its native ego latent using the inferred
controller representation.

For each candidate, we obtain the native rollout from the frozen world model
and extract its ego latent:
\begin{equation}
\begin{aligned}
z_{t+1}^{0,k} &= F_\theta(z_t,a_t^k), \\
\xi_t^k &= P_{\mathrm{ego}}(z_{t+1}^{0,k}).
\end{aligned}
\label{eq:native_ego}
\end{equation}
Here, $\xi_t^k$ denotes the native ego latent for candidate $a_t^k$.

Conditioned on $e_c$, the Latent Bridge predicts a bounded correction to the
native ego latent:
\begin{equation}
\begin{aligned}
u_t^k
    &= B_\phi(\xi_t^k,e_c), \\
\Delta\xi_t^k
    &= g_\phi(\xi_t^k,e_c)\,
       \Pi_{\epsilon,\xi_t^k}(u_t^k).
\end{aligned}
\label{eq:latent_bridge}
\end{equation}
Here, $B_\phi$ proposes a controller-conditioned latent update,
$g_\phi(\cdot)\in[0,1]$ is a learned gate that modulates its contribution,
\begin{equation}
\Pi_{\epsilon,\xi}(u)
=
u\min\!\left\{
1,\,
\frac{\epsilon\|\xi\|_2}
{\max(\|u\|_2,\delta)}
\right\}
\label{eq:bridge_projection}
\end{equation}
projects the update onto an $\ell_2$ trust region of radius
$\epsilon\|\xi\|_2$.

The controller-conditioned rollout is then obtained by injecting the bounded
correction into the ego-related latent:
\begin{equation}
z_{t+1}^{c,k}
=
z_{t+1}^{0,k}
+
\mathcal{I}_{\mathrm{ego}}(\Delta\xi_t^k).
\label{eq:adapted_rollout}
\end{equation}
Here, $\mathcal{I}_{\mathrm{ego}}$ injects the correction only into the
ego-related component of the latent representation.
The remaining components are left unchanged at this adaptation step, and the
adapted state is subsequently propagated through the frozen world-model
dynamics.

%=====这个是普通的表格===========
\begin{table*}[t]
    \centering
    \caption{
   \textbf{Planning performance across frozen world-model planners under the
standard NAVSIM setting.}
For each backbone, \method is applied while keeping its native planning
components frozen.
    }
    \label{tab:standard_navsim}
    \small
    \setlength{\tabcolsep}{4.5pt}
    \begin{tabular}{lcccccccc}
        \toprule
        Method
        & Input
        & \#Traj.
        & Traj. Eval.
        & NC$\uparrow$
        & DAC$\uparrow$
        & EP$\uparrow$
        & TTC$\uparrow$
        & PDMS$\uparrow$ \\
        \midrule
        Human
        & -- & -- & --
        & 100.0 & 100.0 & 87.5 & 100.0 & 94.8 \\

        Constant Velocity
        & -- & 1 & $\times$
        & 69.9 & 58.8 & 49.3 & 49.3 & 21.6 \\

        Ego Status MLP~\cite{NavSim2024}
        & -- & 1 & $\times$
        & 93.0 & 77.3 & 62.8 & 83.6 & 65.6 \\

        VADv2~\cite{Chen2024VADV2}
        & C & 8192 & Rule-based
        & 97.9 & 91.7 & 77.6 & 92.9 & 83.0 \\

        UniAD~\cite{Hu2023PlanningOriented}
        & C & 1 & Rule-based
        & 97.8 & 91.9 & 78.8 & 92.9 & 83.4 \\

        LTF~\cite{Chitta2023PAMI}
        & C & 1 & $\times$
        & 97.4 & 92.8 & 79.0 & 92.4 & 83.8 \\

        PARA-Drive~\cite{weng2024paradrive}
        & C & 1 & Rule-based
        & 97.9 & 92.4 & 79.3 & 93.0 & 84.0 \\

        TransFuser~\cite{Chitta2023PAMI}
        & C \& L & 1 & $\times$
        & 97.7 & 92.8 & 79.2 & 92.8 & 84.0 \\

        LAW~\cite{li2025law}
        & C & 1 & $\times$
        & 96.4 & 95.4 & 81.7 & 88.7 & 84.6 \\

        DRAMA~\cite{yuan2024drama}
        & C \& L & 1 & $\times$
        & 98.0 & 93.1 & 80.1 & 94.8 & 85.5 \\

        Hydra-MDP~\cite{Li2024HydraMDP}
        & C \& L & 8192 & Model-free
        & 98.3 & 96.0 & 78.7 & 94.6 & 86.5 \\

        \wote~\cite{li2025wote}
        & C \& L & 256 & Model-based
        & 98.5 & 96.8 & 81.7 & 94.9 & 88.2 \\

        ResWorld~\cite{zhang2026resworld}
        & C \& L & 256 & Model-based
        & 98.2 & 96.4 & 82.5 & 98.9 & 88.3 \\

        WorldDrive~\cite{gui2026worlddrive}
        & C & 256 & Model-based
        & 98.4 & 96.8 & 83.3 & 95.2 & 89.0 \\

        \midrule

        \wote + \method
        & C \& L & 256 & Model-based
        & 98.5 & 97.3 & 82.0 & 95.2
        & \textbf{88.8}\pdmsgain{0.6} \\
        
        ResWorld + \method
        & C \& L & 256 & Model-based
        & 98.1 & 97.0 & 82.9 & 98.7
        & \textbf{88.7}\pdmsgain{0.4} \\
        
        WorldDrive + \method
        & C & 256 & Model-based
        & 98.4 & 97.0 & 84.0 & 95.1
        & \textbf{89.3}\pdmsgain{0.3} \\

        \bottomrule
    \end{tabular}
\end{table*}
%===========================
\subsection{Bridge Training and Deployment}
\label{sec:bridge_training}

We train the Latent Bridge using controller-conditioned future supervision
together with candidate-level decision supervision. The overall objective is
\begin{equation}
\mathcal{L}_{\mathrm{bridge}}
=
\lambda_{\mathrm{future}}\mathcal{L}_{\mathrm{future}}
+
\lambda_{\mathrm{dec}}\mathcal{L}_{\mathrm{dec}}
+
\lambda_{\mathrm{reg}}\mathcal{L}_{\mathrm{reg}}.
\label{eq:bridge_loss}
\end{equation}

For each candidate, the future loss supervises the adapted BEV prediction
against the future BEV target associated with its controller-specific
execution:
\begin{equation}
\mathcal{L}_{\mathrm{future}}
=
\mathbb{E}_{k,j}
\left[
\ell_{\mathrm{seg}}
\left(
\hat{Y}_{c,j}^{k},
Y_{c,j}^{k}
\right)
\right],
\label{eq:future_loss}
\end{equation}
where $\hat{Y}_{c,j}^{k}$ and $Y_{c,j}^{k}$ denote the predicted and target
future BEV representations of candidate $k$ at rollout step $j$, respectively,
and $\ell_{\mathrm{seg}}$ is implemented using focal loss.

To directly shape candidate preference, we further align the adapted evaluator
with controller-specific candidate quality:
\begin{equation}
\begin{aligned}
p_c &= \operatorname{softmax}(q_c/\tau),\\
\hat{p}_c &= \operatorname{softmax}(\hat{s}_c),\\
\mathcal{L}_{\mathrm{dec}}
&=
-\sum_k p_c^k \log \hat{p}_c^k
+
\mathcal{L}_{\mathrm{aux}}.
\end{aligned}
\label{eq:decision_loss}
\end{equation}
Here, $q_c^k$ is the PDMS score of candidate $k$ under controller $c$,
$\hat{s}_c^k$ is the corresponding score produced by the adapted evaluator,
and $\tau$ is a temperature parameter.
$\mathcal{L}_{\mathrm{aux}}$ collects the additional pairwise-ranking,
metric-level, imitation, and teacher-distillation objectives used during
training.

Finally, we regularize both the applied latent correction and the proposed
update before projection:
\begin{equation}
\mathcal{L}_{\mathrm{reg}}
=
\mathbb{E}_{k,j}
\left[
\|\Delta\xi_j^k\|_2
+
\beta
\left[
\|u_j^k\|_2
-
\epsilon\|\xi_j^k\|_2
\right]_+^2
\right],
\label{eq:regularization_loss}
\end{equation}
where $[x]_+=\max(x,0)$.
The first term discourages unnecessary changes to the native ego latent,
while the second penalizes updates that violate the trust-region bound.
% The first term discourages unnecessary changes to the native ego latent,
% while the second penalizes proposed updates that exceed the prescribed
% trust-region bound.

Training proceeds in two stages.
We first optimize the Controller Encoder using
Eq.~\eqref{eq:execution_loss} and then freeze it.
The Latent Bridge is subsequently optimized using
Eq.~\eqref{eq:bridge_loss}, while the planner, world model, and reward head
remain fixed throughout.
At deployment, the controller representation $e_c$ is computed once from
$\mathcal{S}_c$ and cached; no simulator supervision, online optimization,
or parameter update is required.

\section{Experiments}
\label{sec:experiments}
%===============================

\subsection{Experimental Setup}
\label{sec:exp_setup}

We evaluate \method on NAVSIM~\cite{NavSim2024}, using
\wote~\cite{li2025wote} as the primary frozen world-model planner.
Unless otherwise specified, all comparisons use the same planner, world
model, reward head, and candidate set, isolating the effect of
controller-conditioned adaptation.

We adopt a strict controller-disjoint protocol to evaluate generalization
to unseen deployment controllers. Let $\mathcal{C}_{\mathrm{train}}$ and
$\mathcal{C}_{\mathrm{test}}$ denote the controller sets used for training
and evaluation, respectively, with
$\mathcal{C}_{\mathrm{train}}\cap\mathcal{C}_{\mathrm{test}}=\varnothing$.
All ten evaluation controllers are excluded from both Controller Encoder
pretraining and Latent Bridge training and are observed at test time only
through their support sets $\mathcal{S}_c$.
% For fair comparison, all controller-conditioned variants share the same
% frozen Controller Encoder and support observations.
For each controller, $\mathcal{S}_c$ contains $N=1024$ planned--executed
trajectory pairs. The planned trajectories are drawn from a fixed bank of
representative trajectories obtained by clustering trajectories from the
nuPlan~\cite{Caesar2021nuPlan} training split, and each planned trajectory is paired with its controller-specific execution.

We report the NAVSIM PDMS score as the primary planning metric.
Trajectory ADE measures execution-prediction accuracy, while Spearman's
rank correlation measures the consistency between predicted candidate
ranking and controller-specific candidate quality.
To further examine cases in which controller-induced execution shift has
a larger impact on planning, we report results on a fixed low-score subset
defined by Frozen \wote PDMS $<80$. The subset is determined once from the
frozen baseline and shared across all methods.

%================================================
%========Table1===============
\begin{table}[t]
    \centering
     \caption{
    \textbf{Planning under controller-induced execution shift.}
    Results are averaged over ten controller configurations.
    % Low-score denotes scene--configuration pairs with Frozen \wote PDMS $<80$.
    }
    \label{tab:main_shift}
    \small
    \setlength{\tabcolsep}{4.5pt}
    \begin{tabular}{lcccc}
        \toprule
        & \multicolumn{2}{c}{Full set}
        & \multicolumn{2}{c}{Low-score subset} \\
        \cmidrule(lr){2-3}\cmidrule(lr){4-5}
        Method
        & PDMS $\uparrow$ & $\Delta$
        & PDMS $\uparrow$ & $\Delta$ \\
        \midrule
        Frozen \wote
        & 87.976 & --     & 32.038 & -- \\
        Predicted execution
        & 88.024 & +0.048 & 36.528 & +4.490 \\
        \textbf{\method}
        & \textbf{88.589} & \textbf{+0.613}
        & \textbf{46.128} & \textbf{+14.090} \\
        \bottomrule
    \end{tabular}
\end{table}

% ===========折线图==============
\begin{figure}[t]
    \centering
    \includegraphics[width=\linewidth]{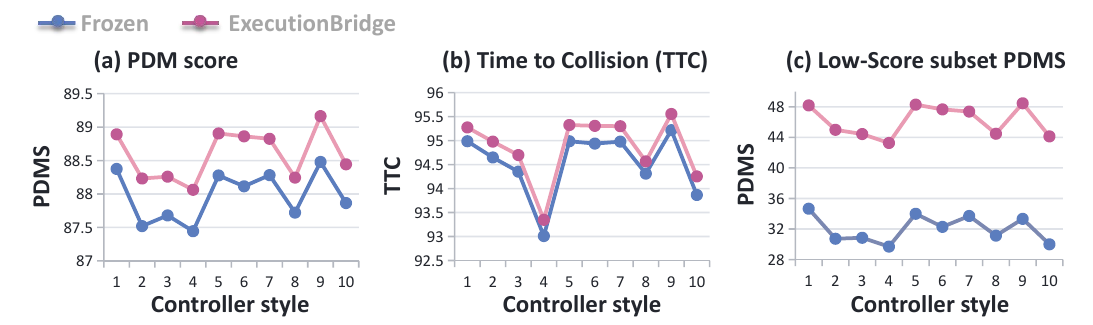}
    \caption{
    \textbf{Planning under unseen controller shifts.}
    Performance of the frozen planner and \method across ten unseen deployment
    controller configurations.
    \textbf{(a)} \method consistently improves PDMS across controller styles
    (mean: $87.98 \rightarrow 88.59$, $+0.61$).
    \textbf{(b)} It also improves TTC
    (mean: $94.53 \rightarrow 94.86$, $+0.33$).
    \textbf{(c)} The improvement is substantially larger on the low-score subset,
    where PDMS increases from $32.04$ to $46.13$ ($+14.09$).
    These results indicate that controller-conditioned adaptation is particularly
    beneficial in challenging driving scenes where candidate evaluation is more
    sensitive to execution differences.
    }
    \label{fig:controller_gain}
\end{figure}

%=====柱状图+仿真图
\begin{figure}[t]
    \centering
    \includegraphics[width=\linewidth]{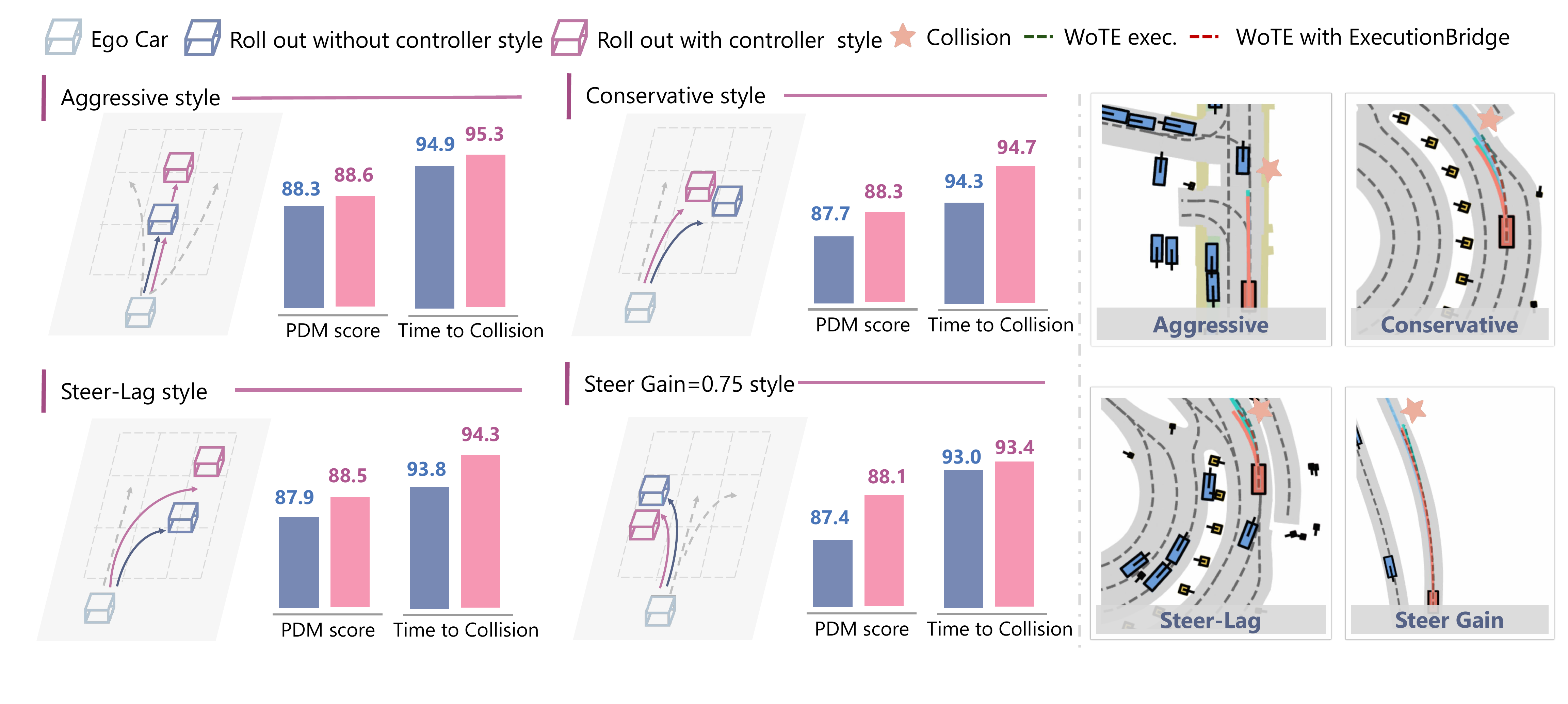}
    \caption{
    \textbf{Adaptation under representative controller styles.}
    We show four representative controller configurations, including aggressive,
    conservative, steering-lag, and steering-gain behaviors.
    Controller-conditioned candidate evaluation consistently improves both PDM
    and Time to Collision (TTC) across these diverse controller styles.
    Examples on the right illustrate safety-critical cases in
    which accounting for controller-specific execution changes the selected
    trajectory and avoids failures under frozen evaluation.
    }
    \label{fig:controller_gain2}
\end{figure}
\begin{figure}[t]
    \centering
    \includegraphics[width=\columnwidth]
    {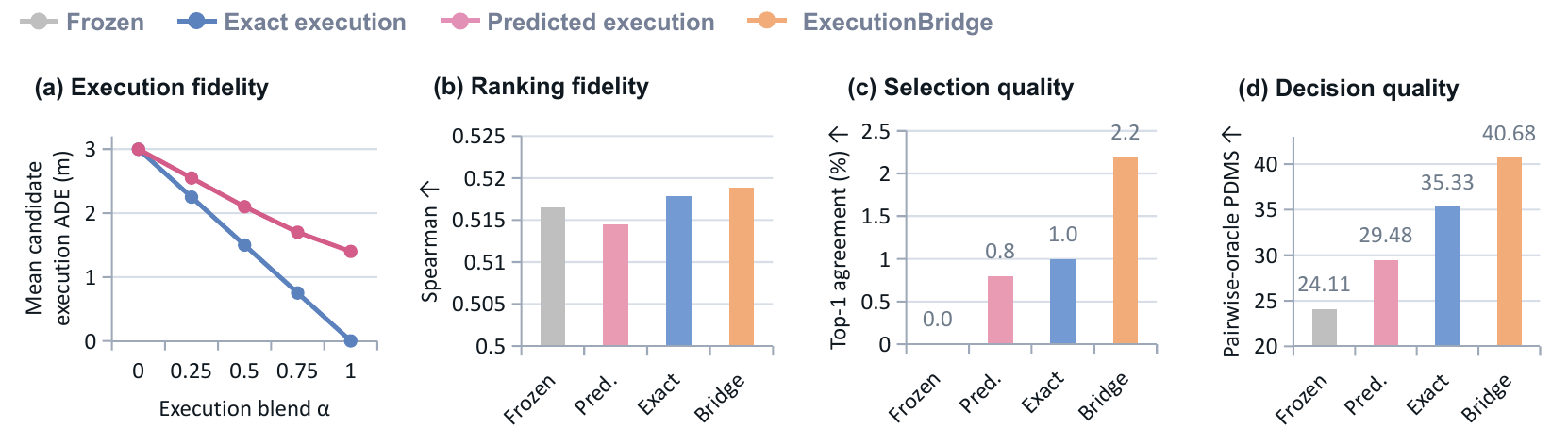}
    \caption{
    \textbf{Execution fidelity does not imply decision fidelity.}
    (a) Predicted and exact controller execution substantially reduce geometric
    execution error.
    (b--c) Nevertheless, candidate-ranking fidelity changes only marginally on
    the frozen-defined difficult-scene subset.
    (d) Direct execution substitution improves selected-candidate quality, but
    ExecutionBridge performs best by preserving the original plan and adapting
    its imagined future.
    Pairwise-oracle PDMS in (d) is a candidate-level diagnostic distinct from
    final navtest PDMS.
    }
    \label{fig:execution_fidelity}
\end{figure}
\subsection{Planning--Execution Gap under Unseen Controllers} \label{sec:exp_shift} 

Across ten held-out controllers, \method consistently improves planning
performance (Table~\ref{tab:main_shift} and Fig.~\ref{fig:controller_gain}).
The improvement is substantially larger on the low-score subset, indicating
that controller-conditioned adaptation is especially beneficial in challenging
scenes where candidate evaluation is more sensitive to execution differences.

Figure~\ref{fig:controller_gain2} illustrates how different controllers
produce different executions of the same planned trajectory through variations
in acceleration, steering response, and temporal lag.
These execution differences can alter the predicted consequence of a candidate
and, consequently, its ranking under the frozen evaluator.
By retaining the planned trajectory as the conditioning input and adapting
the imagined rollout using controller-specific execution information,
\method mitigates such decision-relevant shifts rather than merely improving
trajectory tracking.

\subsection{Execution Fidelity Does Not Imply Decision Fidelity}

We examine whether improving execution accuracy alone is sufficient to improve
candidate evaluation. For each candidate, we interpolate the planned trajectory
toward either its predicted or exact controller execution:
\begin{equation}
a_\alpha^k
=
a^k + \alpha\bigl(\tilde{\tau}^{c,k}-a^k\bigr),
\end{equation}
where $\tilde{\tau}^{c,k}$ denotes either the predicted or exact executed
trajectory and $\alpha\in\{0,0.25,0.5,0.75,1\}$. Thus, $\alpha=0$ recovers the
original plan, while $\alpha=1$ fully substitutes it with the corresponding
execution.

Figure~\ref{fig:execution_fidelity} shows that better execution fidelity
substantially reduces geometric error but only marginally improves candidate
ranking. Although direct execution substitution benefits difficult scenes,
ExecutionBridge yields the best decision quality. This supports adapting the
imagined future while preserving the plan as the world-model
condition.

\subsection{Native Compatibility and Backbone Generality}
\label{sec:exp_standard}
We further evaluate \method under the standard NAVSIM~\cite{NavSim2024} setting and across
different frozen world-model planners.
As shown in Table~\ref{tab:standard_navsim}, \method improves
\wote~\cite{li2025wote} while preserving its native planner, world model,
reward head, and candidate generator.
The same adaptation principle yields consistent gains on
ResWorld~\cite{zhang2026resworld} and
WorldDrive~\cite{gui2026worlddrive}, indicating that the proposed
controller-conditioned adaptation is not tied to a \wote-specific
architecture.
Representative examples in Fig.~\ref{fig:qualitative_standard} further
illustrate how the adapted evaluator leads to more reliable trajectory
selection in challenging driving scenarios.

\begin{table}[t]
    \centering
    \caption{
    \textbf{Ablation of adaptation target and controller information.}
    ``Adapt.'' denotes adaptation of the ego latent $\xi$ or candidate score $s$;
    $e_c$ is the controller representation and $\hat r^{c,k}$ the
    candidate-specific execution residual.
    }
    \label{tab:conditioning_ablation}
    \small
    \setlength{\tabcolsep}{3.0pt}
    \begin{tabular}{@{}lccccc@{}}
        \toprule
        Method
        & Adapt.
        & $e_c$
        & $\hat r^{c,k}$
        & PDMS $\uparrow$
        & Low-score $\uparrow$ \\
        \midrule
        Frozen \wote
        & -- & -- & --
        & 87.976 & 32.038 \\

        Decision-only
        & $\xi$ & -- & --
        & 88.071 & 35.214 \\

        Score calibration
        & $s$ & \checkmark & --
        & 88.184 & 40.327 \\

        Residual-conditioned
        & $\xi$ & \checkmark & \checkmark
        & 88.312 & 45.550 \\

        \textbf{\method}
        & $\xi$ & \checkmark & --
        & \textbf{88.589} & \textbf{46.128} \\
        \bottomrule
    \end{tabular}
\end{table}

%==暂时先不用了
\begin{table}[t]
    \centering
    \caption{
    \textbf{Sensitivity to support-set size.}
    Results over five held-out controller styles show stable execution and
    planning performance across $N=64$--$1024$.
    }
    \label{tab:support_size}
    \small
    \setlength{\tabcolsep}{4.5pt}
    \begin{tabular}{cccc}
        \toprule
        $N$
        & Exec. ADE $\downarrow$
        & Full PDMS $\uparrow$
        & Low-score PDMS $\uparrow$ \\
        \midrule
        64
        & 1.521 & 88.445 & 45.901 \\
        128
        & 1.522 & 88.441 & 45.955 \\
        256
        & 1.521 & 88.441 & 45.954 \\
        1024
        & 1.521 & 88.443 & 45.954 \\
        \bottomrule
    \end{tabular}
\end{table}

\subsection{Ablation and Sensitivity Analysis}
\label{sec:exp_ablation}

\textbf{Adaptation target and controller information.}
Table~\ref{tab:conditioning_ablation} examines whether the gains of
\method arise from additional adaptation capacity, controller-specific
information, or where such information is incorporated.
Results are evaluated over ten held-out controller styles.
Latent adaptation without controller conditioning yields only a modest
improvement over Frozen \wote, indicating that an additional trainable
adapter with decision supervision alone does not explain the gain of
\method.
Conditioning only the final candidate scores on $e_c$ improves performance
further, confirming that the inferred controller representation contains
decision-relevant information.
However, it remains clearly below \method, especially on the low-score
subset, supporting the use of controller context to adapt the imagined
future before evaluation by the frozen reward model.

\textbf{Candidate-specific execution.}
The residual-conditioned variant additionally provides
$\hat r^{c,k}=\hat{\tau}^{c,k}-a^k$ to the Latent Bridge.
Despite this extra candidate-wise information, it performs slightly worse
than conditioning on $e_c$ alone.
This suggests that execution prediction is more useful for identifying
persistent controller characteristics than for providing a direct
per-candidate latent correction.

\textbf{Support-set size.}
Table~\ref{tab:support_size} shows that \method is largely insensitive
to support-set size, with execution ADE and planning performance remaining
nearly unchanged from $N=64$ to $1024$.
This suggests that a small support set is sufficient to infer
controller-specific execution characteristics.

\textbf{Trust-region sensitivity.}
Overly tight trust-region constraints hinder adaptation, while performance
saturates for sufficiently large $\epsilon$; we use $\epsilon=0.2$ throughout.

\section{Conclusion}
\label{sec:conclusion}
We study the \emph{planning--execution gap} in frozen world-model planners,
where deployment controllers can execute the same planned trajectory
differently from the plan-to-motion behavior implicitly captured during
pretraining. We show that execution fidelity does not necessarily imply
decision fidelity: even exact execution provides only limited benefit when
directly substituted for the planned trajectory.

\method addresses this gap by inferring controller-specific execution
characteristics from planned--executed trajectory pairs and adapting the native
ego latent rollout while retaining the original plan as the conditioning input.
It consistently improves planning across unseen controllers and multiple frozen
world-model planners. These results support a simple principle: preserve the plan's semantic role
and use execution evidence to adapt the future imagined for it.

\balance
\bibliographystyle{IEEEtran}
\bibliography{main}
\end{document}